# Concept Relation Discovery and Innovation Enabling Technology (CORDIET)

Jonas Poelmans[1], Paul Elzinga[3], Alexey Neznanov[5], Stijn Viaene[1, 2], Sergei O. Kuznetsov[5], Dmitry Ignatov[5], Guido Dedene[1, 4],

[1]K.U.Leuven, Faculty of Business and Economics, Naamsestraat 69,
3000 Leuven, Belgium
[2]Vlerick Leuven Gent Management School, Vlamingenstraat 83,
3000 Leuven, Belgium
[3]Amsterdam-Amstelland Police, James Wattstraat 84,
1000 CG Amsterdam, The Netherlands
[4]Universiteit van Amsterdam Business School, Roetersstraat 11
1018 WB  Amsterdam, The Netherlands
[5]National Research University Higher School of Economics (HSE), Pokrovskiy boulvard 11
101000 Moscow, Russia
{Skuznetsov, Dignatov, Aneznanov}@hse.ru
{Jonas.Poelmans, Stijn.Viaene, Guido.Dedene}@econ.kuleuven.be
Paul.Elzinga@amsterdam.politie.nl

**Abstract.** Concept Relation Discovery and Innovation Enabling Technology (CORDIET), is a toolbox for gaining new knowledge from unstructured text data. At the core of CORDIET is the C-K theory which captures the essential elements of innovation. The tool uses Formal Concept Analysis (FCA), Emergent Self Organizing Maps (ESOM) and Hidden Markov Models (HMM) as main artifacts in the analysis process. The user can define temporal, text mining and compound attributes. The text mining attributes are used to analyze the unstructured text in documents, the temporal attributes use these document's timestamps for analysis. The compound attributes are XML rules based on text mining and temporal attributes. The user can cluster objects with object-cluster rules and can chop the data in pieces with segmentation rules. The artifacts are optimized for efficient data analysis; object labels in the FCA lattice and ESOM map contain an URL on which the user can click to open the selected document.

## 1. Introduction

In many law enforcement organizations, more than 80 % of available data is in textual form. In the Netherlands and in particular the police region Amsterdam-Amstelland the majority of these documents are observational reports describing observations made by police officers on the street, during motor vehicle inspections, police patrols, interventions, etc. Intelligence Led Policing (ILP) aims at making the shift from a traditional reactive intuition-led style of policing to a proactive intelligence led approach (Collier 2006). Whereas traditional ILP projects are typically based on statistical analysis of structured data, e.g. geographical profiling of street robberies, we go further by uncovering the underexploited potential of unstructured textual data.

In this paper we report on our ongoing research projects on concept discovery in law enforcement and the CORDIET tool that is being developed based on this research. At the core of CORDIET is the Concept-Knowledge (C-K) theory (Poelmans et al. 2009) which structures the KDD process. For each of the 4 transitions in the design square functionality is provided to support the data analyst or domain expert in exploring the data. First, the data source and the ontology containing the attributes used to analyze these data files should be loaded into CORDIET. In the ontology, the user can define temporal, text mining and compound attributes. The text mining attributes are used to analyze the unstructured text in documents, the temporal attributes use these document's timestamps for analysis. The compound attributes are XML rules based on text mining and temporal attributes. The user can cluster objects with object-cluster rules and can chop the data in pieces with segmentation rules. After the user selected the relevant attributes, rules and objects, the analysis artifacts can be created. The tool can be used to create FCA lattices, ESOMs and HMMs. The artifacts are optimized for efficient data analysis; object labels in the FCA lattice and ESOM map contain an URL on which the user can click to open the selected document. Afterwards the knowledge products such as a 27-construction for a human trafficking suspect can be deployed to the organization.

## 2. Data analysis artifacts

In this section we briefly describe the data analysis and visualisations artifacts that can be created with the CORDIET software. The tool uses Formal Concept Analysis (FCA), Emergent Self Organizing Maps (ESOM) and Hidden Markov Models (HMM) as main artifacts in the analysis process.

### 2.1 Formal Concept Analysis

Formal Concept Analysis (FCA), a mathematical unsupervised clustering technique originally invented by Wille (1982) offers a formalization of conceptual thinking. The intuitive visualization of concept lattices derived from formal contexts has had many applications in the knowledge discovery field (Stumme et al. (1998), Poelmans et al. (2010b)). Concept discovery is an emerging discipline in which FCA based methods are used to gain insight into the underlying concepts of the data. In contrast to standard black-box data mining techniques, concept discovery allows analyzing and refining these underlying concepts and strongly engages the human expert in the data discovery exercise. The main goal is to make previously inaccessible information available for practitioners easy to interpret visual display. In particular, the visualization capabilities are of interest to the domain expert who wants to explore the information available, but at the same time has not much experience in mathematics or computer science. The details of FCA theory and how we used it for KDD can be found in (Poelmans et al. 2009). Traditional FCA is mainly using data attributes for concept analysis. We also used process activities (events) as attributes (Poelmans et al. 2010c). Typically, coherent data attributes were clustered to reduce the computational complexity of FCA.

## 2.2 Temporal Concept Analysis

Temporal Concept Analysis (TCA) is an extension of traditional FCA that was introduced in scientific literature about nine years ago (Wolff 2005). TCA addresses the problem of conceptually representing time and is particularly suited for the visual representation of discrete temporal phenomena. The pivotal notion of TCA theory is that of a conceptual time system. In the visualization of the data, we express the "natural temporal ordering" of the observations using a time relation R on the set G of time granules of a conceptual time system. We also use the notions of transitions and life tracks. The basic idea of a transition is a "step from one point to another" and a life track is a sequence of transitions (Elzinga et al. 2010).

## 2.3 Emergent Self Organising Maps

Emergent Self Organizing Maps (ESOM) (Ultsch 2003) are a special class of topographic maps. ESOM is argued to be especially useful for visualizing sparse, high-dimensional datasets, yielding an intuitive overview of its structure. Topographic maps perform a non-linear mapping of the high-dimensional data space to a low-dimensional one, usually a two-dimensional space, which enables the visualization and exploration of the data. ESOM is a more recent type of topographic map. According to Ultsch, "emergence is the ability of a system to produce a phenomenon on a new, higher level". In order to achieve emergence, the existence and cooperation of a large number of elementary processes is necessary. An emergent SOM differs from a traditional SOM in that a very large number of neurons (at least a few thousands) are used (Ultsch et al. 2005). In the traditional SOM, the number of nodes is too small to show emergence.

## 2.4 Hidden Markov Models

A Hidden Markov Model (HMM) is a statistical technique that can be used to classify and generate time series. A HMM (Rabiner 1989) can be described as a quintuplet I = (A, B, T, N, M), where N is the number of hidden states and A defines the probabilities of making a transition from one hidden state to another. M is the number of observation symbols, which in our case are the activities that have been performed to the patients. B defines a probability distribution over all observation symbols for each state. T is the initial state distribution accounting for the probability of being in one state at time t = 0. For process discovery purposes, HMMs can be used with one observation symbol per state. Since the same symbol may appear in several states, the Markov model is indeed "hidden". We visualize HMMs by using a graph, where nodes represent the hidden states and the edges represent the transition probabilities. The nodes are labelled according to the observation symbol probability.

## 3. Data sources

In this research three main data sources have been used. The first data source was the police database "Basis Voorziening Handhaving" (BVH) of the Amsterdam-Amstelland police. Multiple datasets were extracted from this data source, including the domestic violence, human trafficking and terrorism dataset. The second data source was the World Wide Web, from which we collected over 700 scientific articles. The third dataset consist of 148 breast cancer patients that were hospitalized during the period from January 2008 till June 2008.

### 3.1 Data source BVH

The database system BVH is used by all police forces of the Netherlands and the military police, the Royal Marechaussee. This database system contains both structured and unstructured textual information. The contents of the database are subdivided in two categories: incidents and activities. Incident reports describe events that took place that are in violation with the law. These include violence, environmental and financial crimes. During our research we analyzed the incident reports describing violent incidents and we aimed at automatically recognizing the domestic violence cases.

Activities are often performed after certain incidents occurred and include interrogations, arrestment, etc., but activities can also be performed independent of any incident, such as motor vehicle inspections, an observation made by a police officer of a suspicious situation, etc. Each of these activities performed are described in a textual report by the responsible officer. We used the observations made by police officers to find indications for human trafficking and radicalizing behavior.

In the year 2005, Intelligence Led Policing was introduced at the police of Amsterdam, resulting in a sharp increase in the number of filed activity reports describing observations made by police officers, i.e. from 34817 in 2005 to 67584 in 2009. These observational reports contain a short textual description of what has been observed and may be of great importance for finding new criminals. The involved persons and vehicles are stored in structured data fields in a separate database table and are linked to the unstructured report in a separate database table using relational tables. The content of all these database tables is then used by the police officer to create a document containing all the information. We however did not use these generated documents because it is possible that the information in the database tables is modified afterwards without updating the generated documents.

Therefore, we wrote an export program that automatically composes documents based on the most recent available information in the databases. These documents are stored in XML format and can be read by the CORDIET toolset.

Before our research, no automated analyses were performed on the observational reports written by officers. The reason was an absence of good instruments to detect the observations containing interesting information and to analyze the texts they contain. Only on the structured information stored in police databases, analyses were performed. These include the creation of management summaries using Cognos information cubes, geographical analysis of incidents with Polstat and data mining with Datadetective.

### 3.2 Data source scientific articles

For the survey of FCA research articles, we used the CORDIET toolset. Over 700 pdf files containing articles about FCA research were downloaded from the WWW and automatically analyzed. The structure of the majority of these papers was as follows:

1. Title of the paper

2. Author names, addresses, emails
   3. Abstract and keywords
   4. The contents of the article
   5. The references

During our research we used parts 1, 2 and 3. Parts 2 and 3 to detect the research topics covered in the papers. Part 1 was used for doing a social analysis on the authors of the papers i.e. which research groups are working on which topics, etc.

During the analysis, these pdf-files were converted to ordinary text and the abstract, title and keywords were extracted. The open source tool Lucene was used to index the extracted parts of the papers using the thesaurus. The result was a cross table describing the relationships between the papers and the term clusters or research topics from the thesaurus. This cross table was used as a basis to generate the lattices.

We only used abstract, title and keywords because the full text of the paper may mention a number of concepts that are irrelevant to the paper. For example, if the author who wrote an article on information retrieval gives an overview of related work mentioning papers on fuzzy FCA, rough FCA, etc., these concepts may be irrelevant however they are detected in the paper. If they are relevant to the entire paper we found they were typically also mentioned in title, abstract or keywords.

One of the central components of our text analysis environment is the thesaurus containing the collection of terms describing the different research topics. The initial thesaurus was constructed based on expert prior knowledge and was incrementally improved by analyzing the concept gaps and anomalies in the resulting lattices. The thesaurus is a layered thesaurus containing multiple abstraction levels. The first and finest level of granularity contains the search terms of which most are grouped together based on their semantic meaning to form the term clusters at the second level of granularity.

The term cluster "Knowledge discovery" contains search terms "data mining", "KDD", "data exploration", etc. which can be used to automatically detect the presence or absence of the "Knowledge discovery" concept in the papers. Each of these search terms were thoroughly analyzed for being sufficiently specific. For example, we first had the search term "exploration" for referring to the "Knowledge Discovery" concept, however when we used this term we found that it also referred to concepts such as "attribute exploration" etc. Therefore we only used the specific variant such as "data exploration", which always refers to the "Knowledge Discovery" concept. We aimed at composing term clusters that are complete, i.e. we searched for all terms typically referring to for example the "information retrieval" concept. Both specificity and completeness of search terms and term clusters was analyzed and validated with FCA lattices on our dataset.

### 3.3 Data source clinical pathways

The third dataset consist of 148 breast cancer patients that were hospitalized during the period from January 2008 till June 2008. They all followed the care trajectory determined by the clinical pathway Primary Operable Breast Cancer (POBC), which structures one of the most complex care processes in the hospital. Before the patient is hospitalized, she ambulatory receives a number of pre-operative investigative tests. During the surgery support phase she is prepared for the surgery she will receive, while being in the hospital. After surgery she remains hospitalized for a couple of days until she can safely go home. The post-operative activities are also performed in an ambulatory fashion. Every activity or treatment step performed to a patient is logged in a database and in the dataset we included all the activities performed during the surgery support phase to each of these patients. Each activity has a unique identifier and we have 469 identifiers in total for the clinical path POBC. Using the timestamps assigned to the performed activities, we turned the data for each patient into a sequence of events. These sequences of events were used as input for the process discovery methods. We also clustered activities with a similar semantic meaning to reduce the complexity of the lattices and process models. The resulting dataset is a collection of XML files where each XML corresponds with exactly one activity.

## 4. CORDIET system architecture and business use case diagram

### 4.1 Business use case diagram

In Poelmans et al. (2010) we instantiated the C- K design theory with FCA and ESOM and showed it was an ideal framework to structure the KDD process on a conceptual level as multiple iterations through a design square. The C-K theory is also at the core of CORDIET. For each C-K phase there are use cases that describe the functionality of the phases. The results of the use cases of a previous phase serve as input for the use cases of the next phases. The business use case model in Figure 1clearly shows this C-K inspired architecture of CORDIET.

The first C-K space, "start investigation", aims at transforming existing knowledge and information into objects, attributes, ontology elements etc. (conceptualization). The second C-K phase, "compose artifact", will create artifacts from the data that visualize its underlying concepts and conceptual relationships (concept expansion). The third C-K phase, "analyze artifact", is about distilling new knowledge from these concept representations. The fourth and last C-K phase is about summarizing this newly gained knowledge and feeding it back to the domain experts who can incorporate it in their way of working. After this final step, a new C-K iteration can start based on the original information and/or newly added knowledge. Iterating though the design square will stop when no new knowledge can be found anymore.

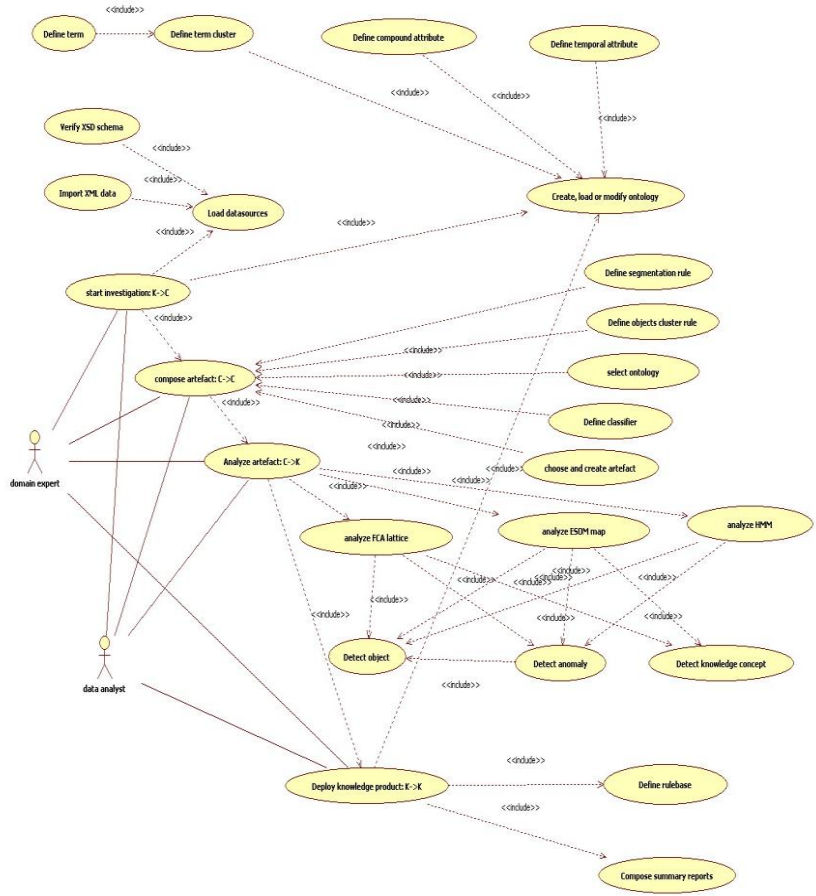

**Fig. 1.** Business use case diagram of CORDIET.

### 4.2 The software lifecycles of CORDIET

The architecture of the CORDIET software underwent some serious changes during the development of this research. During the first stage of this research we were working on the domestic violence data (Poelmans et al. 2010) and CORDIET consisted of an FCA, ESOM component and a commercial text mining tool was used to index the documents. Our own programming took care of the documents extraction from the database and the conversion of the data to be used as input for the artefact creation components. This first version had its limitations and was seriously modified for the terrorism and human trafficking research. Amongst others, indexing of documents was done with Lucene.

A separate RDBMS database was used for the maintenance of the ontology with an ERD model. The latest version used a topic map for maintaining the ontology and the open source topic map editor "ontopoly". This latest version will be described in detail in this chapter.

### 4.3 The development of an operational version of CORDIET

KULeuven and Moscow Higher School of Economics decided to jointly develop an operational software system based on the latest version of CORDIET toolbox. This system will be a user friendly application making visualizations such as FCA, ESOM and HMM available to its users. This version of the toolbox will be based on a distributed web service architecture. Web services are a well standardized, easy to access and flexible piece of technology that can be adapted for different languages and environments.

As a consequence, all input/output activities are represented as XML. Figure 2 shows the general architecture of the new version of CORDIET.

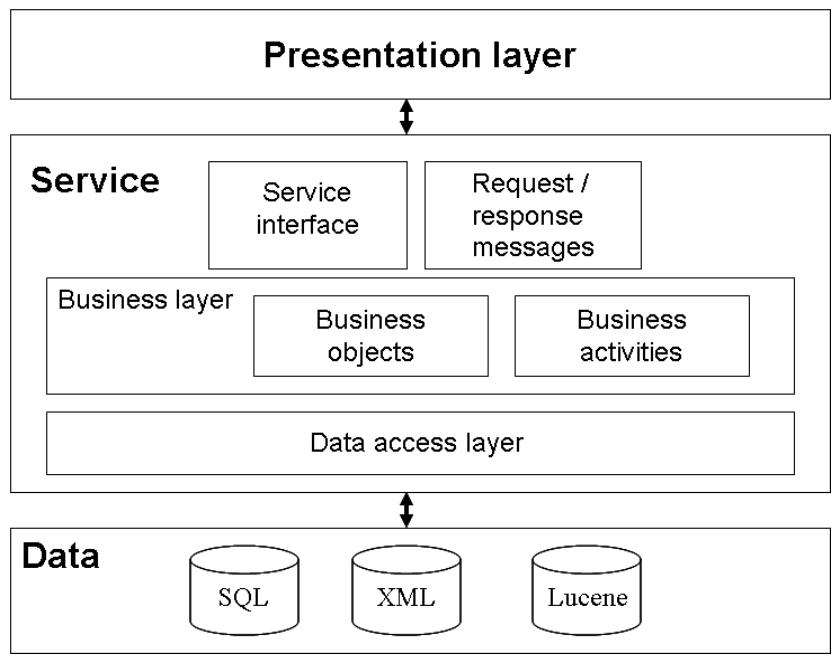

**Fig. 2.** A representation of the CORDIET web service oriented architecture

### 4.3.1    Presentation layer

The presentation layer is the graphical user interface where the interactions of the user with the system are handled. This presentation layer is being developed using Silverlight. CORDIET will use two types of main visualizations. The master mode will mainly be used by domain experts who have limited knowledge of data analysis. The user will be able to load a profile for each of the four C-K transition steps, this profile contains all information the tool needs to automatically complete the step in the data analysis. This profile has been prepared by a data analyst. The user can go to the advanced mode. In the advanced mode, he can fully edit an existing or create a new profile. In the advanced mode, a graph-like display will be used to create, modify and compose different attributes.

#### 4.3.2 Service layer

The service layer will be the core of CORDIET. The service interface takes care of the I/O activities with the presentation layer and accessing of the data through the data access layer.

#### 4.3.3 Business layer

The business layer is divided into two sections, the business objects and the business activities which refer to the different activities within the C/K cycle. This layer includes functionality for creating HMMs, ESOMs, Concept lattices, etc.

#### 4.3.4 Data access layer

The data access layer is used to access the data sections: the relational database, the XML data and the Lucene indexes. The data sets consist of a relational database (PostgresSQL), a dataset with XML files and a Lucene index. The data-indexer component reads the XML files from a selected dataset, parses the XML into the SQL database and generates the Lucene index.

#### 4.3.5 Language module

Different languages including English, Dutch and Russian should be supported. The user must be able to choose between these languages. The version of Lucene indexer of documents used has a large variety of analyzers like Russian Analyzer, Dutch Analyzer, German Analyzer etc. The default Analyzer is English.

## 5. Conclusions

In this paper we briefly described the toolbox, Concept Relation Discovery and Innovation Enabling Technology (CORDIET), for gaining new knowledge from unstructured text data. This toolbox has been embedded within the C-K theory, which captures the essential elements of innovation. The tool uses Formal Concept Analysis (FCA), Emergent Self Organizing Maps (ESOM) and Hidden Markov Models (HMM) as main artifacts in the analysis process.

At the core of the CORDIET toolbox is the business use case where the C-K transitions are mapped on the functionalities of the toolbox. The C-K functionalities are described in detail and demonstrated with real life cases. CORDIET in its current version has become a very powerful toolbox for mining all general reports of the BVH database of the past 5 years. KULeuven and Moscow Higher School of Economics decided to jointly develop an operational software system based on the latest version of CORDIET toolbox. This system will be a user friendly application making visualizations such as FCA, ESOM and HMM more uniformly available to its users where as the current toolbox makes use of integrated open source packages with different user interfaces.

**Acknowledgements**
Jonas Poelmans is aspirant of the "Fonds voor Wetenschappelijk Onderzoek – Vlaanderen" or "Research Foundation Flanders".

# References

[1] Collier, P.M. (2006) Policing and the intelligent application of knowledge. Public money & management. Vol. 26, No. 2, pp. 109-116.

[2] Elzinga, P., Poelmans, J., Viaene, S., Dedene, G., Morsing, S. (2010) Terrorist threat assessment with Formal Concept Analysis. Proc. IEEE International Conference on Intelligence and Security Informatics. May 23-26, 2010 Vancouver, Canada. ISBN 978-1-42446460-9/10, 77-82.

[3] Poelmans, J., Dedene, G., Verheyden, G., Van der Mussele, H., Viaene, S., Peters, E. (2010c). Combining business process and data discovery techniques for analyzing and improving integrated care pathways. Lecture Notes in Computer Science, Advances in Data Mining. Applications and Theoretical Aspects, 10th Industrial Conference (ICDM), Leipzig, Germany, July 12-14, 2010. Springer

[4] Poelmans, J., Elzinga, P., Viaene, S., Dedene, G. (2009). A case of using formal concept analysis in combination with emergent self organizing maps for detecting domestic violence. In : Lecture Notes in Artificial Intelligence, Vol. 5633(XI), (Perner, P. (Eds.)). Industrial conference on data mining ICDM 2009. Leipzig (Germany), 20-22 July 2009 (pp. 402 p.).

[5] Poelmans, J., Elzinga, P., Viaene, S., Dedene, G. (2010b), Formal Concept Analysis in knowledge discovery: a survey. Lecture Notes in Computer Science, 6208, 139-153, 18th international conference on conceptual structures (ICCS 2010): from information to intelligence. 26 - 30 July, Kuching, Sarawak, Malaysia. Springer.

[6] Rabiner, L.R. (1989) A tutorial on Hidden Markov Models and selected applications in speech recognition. Proceedings IEEE 77 (2): 257-286.

[7] Stumme, G., Wille, R., Wille, U. (1998) Conceptual knowledge discovery in databases using Formal Concept Analysis Methods. PKDD, 450-458.

[8] Ultsch, A. (2003) Maps for visualization of high-dimensional Data Spaces. In proc. WSOM'03, Kyushu, Japan, pp. 225-230.

[9] Ultsch, A., Hermann, L. (2005) Architecture of emergent self-organizing maps to reduce projection errors. In Proc. ESANN 2005, pp. 1-6.

[10] Wille, R. (1982). Restructuring lattice theory: an approach based on hierarchies of concepts. I. Rival (Ed.): Ordered sets, 445-470. Reidel. Dordrecht-Boston.

[11] Wolff, K.E. (2005) States, transitions and life tracks in Temporal Concept Analysis. In: B. Ganter et al. (Eds.): Formal Concept Analysis, LNAI 3626, pp. 127-148. Springer, Heidelberg.